\documentclass[11pt]{article}

\usepackage[final]{acl}

\usepackage{times}
\usepackage{latexsym}

\usepackage[T1]{fontenc}
\usepackage[utf8]{inputenc}

\usepackage{microtype}

\usepackage{inconsolata}

\usepackage{graphicx}
\usepackage{geometry}
\usepackage{algorithm}
\usepackage{algpseudocode}
\usepackage{amsmath}
\usepackage{booktabs}
\usepackage{amsmath}
\usepackage{multirow}
\usepackage{pifont}
\usepackage{tcolorbox}
\usepackage{verbatim}
\tcbuselibrary{breakable}

\usepackage{colortbl}
\usepackage{multirow}
\usepackage{multicol}
\usepackage{makecell}
\usepackage{tabularx}
\usepackage[table]{xcolor} 
\definecolor{bgmark}{gray}{0.9} 
\usepackage{enumitem}

\usepackage{xspace}  
\newcommand{\sys}{\texttt{CauScientist}\xspace}
\definecolor{morandiorange}{HTML}{FDE6D0}
\title{\sys: Teaching LLMs to Respect Data for Causal Discovery}

\author{Bo Peng$^{1,2,3}$, Sirui Chen$^{1,4}$, Lei Xu$^{1,5}$, Chaochao Lu$^{1}$\footnotemark[2] \\
  $^{1}$Shanghai Artificial Intelligence Laboratory,~~~$^{2}$Shanghai Jiao Tong University \\
  $^{3}$Shanghai Innovation Institute~~~$^{4}$Tongji University~~~
  $^{5}$École Polytechnique Fédérale de Lausanne\\
  \texttt{peng\_bo2019@sjtu.edu.cn}, 
  \texttt{luchaochao@pjlab.org.cn}
  }

\begin{document}
\maketitle
\begin{abstract}
Causal discovery is fundamental to scientific understanding and reliable decision-making. Existing approaches face critical limitations: purely data-driven methods suffer from statistical indistinguishability and modeling assumptions, while recent LLM-based methods either ignore statistical evidence or incorporate unverified priors that can mislead result. To this end, we propose \sys, a collaborative framework that synergizes LLMs as hypothesis-generating ``data scientists'' with probabilistic statistics as rigorous ``verifiers''. \sys employs hybrid initialization to select superior starting graphs, iteratively refines structures through LLM-proposed modifications validated by statistical criteria, and maintains error memory to guide efficient search space. Experiments demonstrate that \sys substantially outperforms purely data-driven baselines, achieving up to 53.8\% F1 score improvement and enhancing recall from 35.0\% to 100.0\%. Notably, while standalone LLM performance degrades with graph complexity, \sys reduces structural hamming distance (SHD) by 44.0\% compared to Qwen3-32B on 37-node graphs. Our project page is at \url{https://github.com/OpenCausaLab/CauScientist}.

\end{abstract}
\renewcommand*{\thefootnote}{\fnsymbol{footnote}}
% \footnotetext[1]{Equal contribution.}
\footnotetext[2]{Corresponding author.}
\section{Introduction}
\begin{figure}[t]
    \centering
    \includegraphics[width=.95\linewidth]{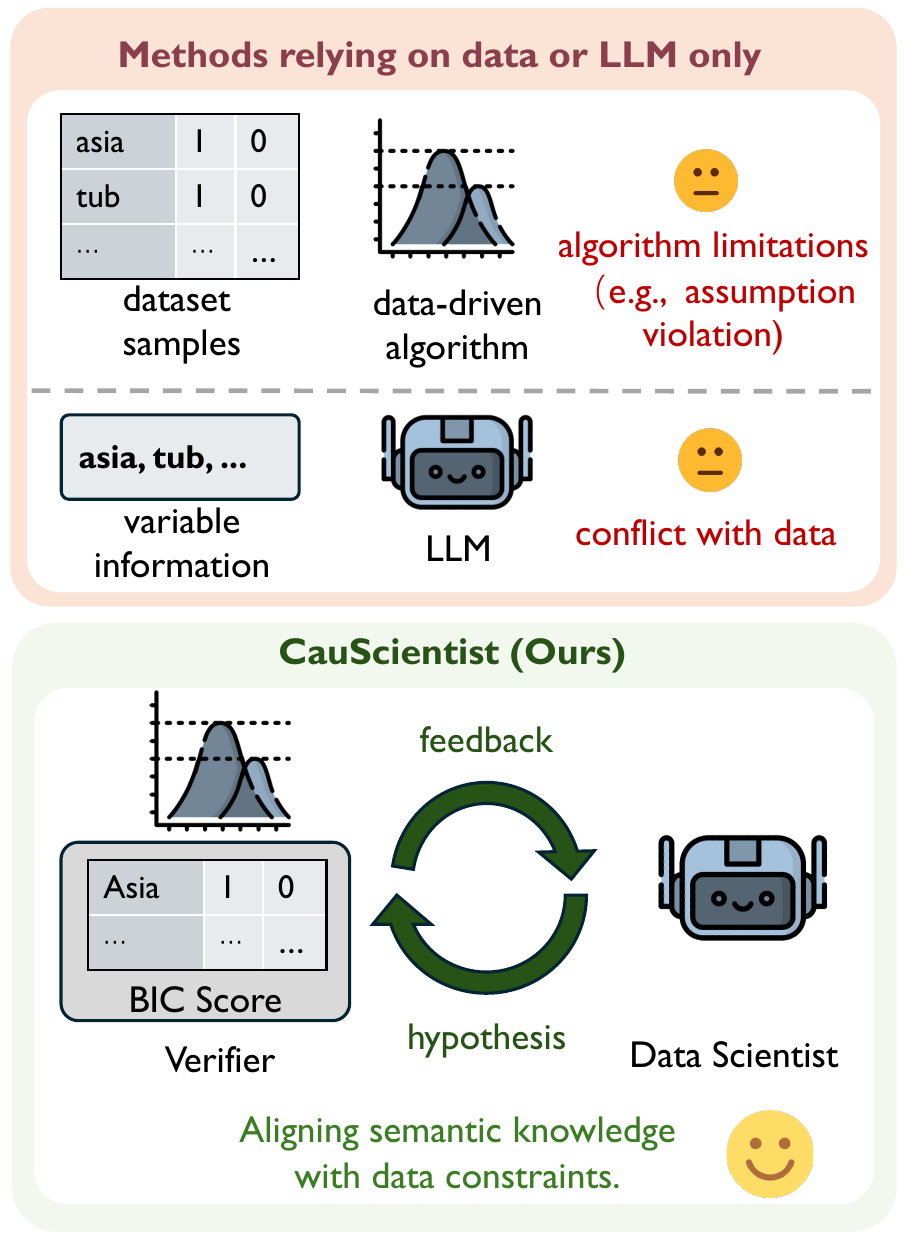} 
    \caption{Conceptual comparison of causal discovery methods. Data-driven models produce data-faithful answers but with inherent algorithm limitations. LLMs generate logically plausible answers, yet they often contradict statistical regularities. \sys combines the strengths of both methods, aligning semantic knowledge with data constraints.}
    \label{fig:llm_vs_stat}
\end{figure}
Causal discovery~\citep{spirtes2000causation,pearl2009causality}, the inference of causal structure from observational data, serves as a cornerstone for scientific inquiry and robust artificial intelligence.
While purely data-driven methods have progressed from discrete constraint-based search (e.g., FCI~\cite{10.5555/2074158.2074215}) to continuous optimization and amortized inference (e.g., NOTEARS~\citet{zheng2018dags}, AVICI~\cite{lorch2022amortized}), they remain fundamentally limited by statistical indistinguishability (e.g., equivalence classes), non-convex objectives and modeling assumptions, and sensitivity to distribution shift.

To overcome these statistical limitations, using large language models (LLMs) for causal discovery has emerged as a promising direction. Trained on vast corpora containing explicit causal statements (e.g., ``stock prices fell due to an interest-rate hike'' in news reports), LLMs acquire rich causal knowledge and the ability to infer causal relationships from semantic information.
Currently, there are two predominant paradigms for LLM-assisted causal discovery. The first involves leveraging LLMs to construct causal graphs directly from semantic information \cite{jiralerspong2024efficient, roy2025causal}. However, this approach fails to fully utilize statistical data, potentially yielding causal relations that conflict with empirical distributions. The second paradigm uses LLMs to provide prior knowledge that informs traditional data-driven methods \cite{long2023causal, takayama2025integrating}. Nevertheless, these methods often lack mechanisms to validate the correctness of LLM-derived priors, allowing erroneous information to compromise subsequent statistical estimation. Therefore, how to effectively integrate semantic information with statistical methods remains a pivotal yet unresolved problem.

To bridge this gap, we propose \sys, a collaborative framework that integrates LLM as ``data scientist'' with probabilistic statistic serving as ``verifier''. Figure \ref{fig:llm_vs_stat} visualizes how our framework integrates knowledge and data: unlike methods relying on a single source of evidence, \sys aligns semantic knowledge with data constraints through iterative verification. Specifically, \sys operates through the following stages:
(1) \textbf{Hybrid initialization}. Without assuming any fixed priors, \sys first generates candidate causal graphs from both standard data-driven algorithms and LLM. The graph with the superior Bayesian Information Criterion (BIC) \citep{schwarz1978estimating} is selected as the initial graph. BIC serves as a statistical metric that evaluates the trade-off between data fidelity (how well the graph explains the data) and structural complexity (the number of edges or parameters).
(2) \textbf{Collaborative verification and refinement}. The LLM proposes structural modifications to the initial graph, which are subsequently scrutinized by a verifier. To optimize the search, rejected modifications are logged in an error memory to guide the LLM in pruning the search space and preventing redundant errors in successive rounds.
(3) \textbf{Iterative optimization}. This refinement process is iterated until convergence to the final graph.
Overall, \sys strikes a balance between the LLM's rich background knowledge and the rigorous constraints of statistical methods, enabling effective and robust causal discovery.

We conduct extensive experiments on datasets with varying graph scales and causal relationships. The results show that \sys substantially outperforms purely data-driven baselines, achieving up to a 53.8\% gain in F1 score and improving recall from 35.0\% to 100.0\% in the best case. We further identify a key limitation of relying solely on LLMs for causal discovery: performance degrades markedly as graph size increases. In contrast, \sys reduces SHD by 44.0\% relative to Qwen3-32B on graph with 37 nodes, yielding more reliable causal relationships.

To summarize, our main contributions are:
\begin{itemize}[itemsep=2pt,topsep=2pt,parsep=0pt,leftmargin=*]
    \item We propose \sys, a collaborative causal discovery framework that leverages the LLM’s rich background knowledge while enforcing the rigorous constraints of statistical methods.
    \item We instantiate this collaboration with a BIC-based verifier that evaluates proposed structural modifications, and an error memory that guides the LLM to prune the search space efficiently and avoid redundant proposals.
    \item We conduct comprehensive experiments to validate the effectiveness of \sys, and demonstrate its generality and robustness on varying graph scales and causal relationships.
\end{itemize}

\section{Related Work}
\begin{table*}[t]
\centering
\small
% 自动缩放表格以适应页面宽度
\resizebox{\textwidth}{!}{%
\setlength{\tabcolsep}{4pt}
\begin{tabular}{lcccc}
\toprule
\textbf{Method} & \textbf{LLM Role} & \textbf{Error Feedback} & \textbf{Statistical Prior} & \textbf{Verification} \\
% \midrule
% \multicolumn{5}{l}{\textit{Category I: Direct Structure Inference}} \\
\hline
\multicolumn{5}{c}{\cellcolor{green!10}\emph{Category I: Direct Structure Inference}}\\
\hline
LLM-BFS \citep{jiralerspong2024efficient} & Search Agent & Unidirectional & \ding{55} & \ding{55} \\
Causal-LLM \citep{roy2025causal} & Generator & Unidirectional & \ding{55} & \ding{55} \\
% \midrule
% \multicolumn{5}{l}{\textit{Category II: Knowledge Injection \& Priors}} \\
\hline
\multicolumn{5}{c}{\cellcolor{teal!10}\emph{Category II: Knowledge Injection \& Priors}}\\
\hline
SCP \citep{takayama2025integrating} & Prior & Unidirectional & \ding{51} & \ding{55} \\
Causal Order \citep{vashishtha2025causal} & Prior & Unidirectional & \ding{55} & \ding{55} \\
ET-MCMC \citep{10966043} & Prior & Unidirectional & \ding{55} & Soft Penalty \\
LLM-MEC \citep{long2023causal} & Prior & Unidirectional & \ding{51} & MEC Consistency \\
% \midrule
% \multicolumn{5}{l}{\textit{Category III: Iterative Co-Refinement}} \\
\hline
\multicolumn{5}{c}{\cellcolor{orange!10}\emph{Category III: Iterative Co-Refinement}}\\
\hline
CMA \citep{abdulaal2023causal} & Refiner & Bi-directional & \ding{55} & \ding{55} \\
\textbf{\sys (Ours)} & \textbf{Refiner} & \textbf{Bi-directional} & \ding{51} & BIC Score \\
\bottomrule
\end{tabular}%
}
\caption{Comparison of LLM-integrated causal discovery paradigms. \textit{Direct Inference} methods use LLMs to construct graphs primarily via metadata. \textit{Knowledge Injection} methods use LLMs to constrain statistical algorithms. \textit{Iterative Co-Refinement} methods enable bi-directional optimization.}
\label{tab:comparison}
\end{table*}

\label{sec:related_work}
We categorize these approaches based on the structural role of the LLM: from direct inference to auxiliary injection, and finally to iterative co-refinement. Table~\ref{tab:comparison} summarizes the comparison between our method and prior works.

\subsection{Direct Structure Inference}
In this paradigm, the LLM acts as the primary engine to construct the causal graph directly from metadata, treating discovery as a generation or search problem.
\citet{jiralerspong2024efficient} employ LLMs to guide a Breadth-First Search (BFS) over the graph space. While they utilize statistical tests to prune the search, the interaction is limited: the LLM functions as a generator, and the data acts merely as a passive filter. 
\citet{roy2025causal} propose Causal-LLM, a unified framework offering two modes: a prompt-based method and a data-driven approach. However, the modes in Causal-LLM operate disjointly—the prompt-based mode relies purely on LLM capabilities without data constraints, while the data-driven mode ignores textual knowledge. The direct structure inference methods lack a recovery mechanism: once a proposal is pruned, the agent does not receive feedback to correct its strategy.

\subsection{Knowledge Injection and Priors}
A more rigorous paradigm integrates LLM knowledge as \textbf{auxiliary signals} to guide or constrain standard statistical algorithms. \citet{takayama2025integrating} propose "Statistical Causal Prompting" (SCP), utilizing LLM outputs as initialization priors for algorithms like PC. Similarly, \citet{vashishtha2025causal} infer causal ordering priors, and \citet{10966043} integrate soft ancestral constraints within a Bayesian framework. \citet{long2023causal} address the identifiability limit by using LLMs to orient undirected edges within a Markov Equivalence Class (MEC). Despite their utility, these integrations remain unidirectional. There is typically no feedback loop for the statistical model to correct the LLM's false beliefs when they conflict with observed data. Furthermore, methods like \citet{long2023causal} operate under a fixed-skeleton assumption: they are restricted to orienting a pre-computed graph and lack the capacity to correct global structural errors (e.g., missing or spurious edges) inherent in the initial MEC.

\subsection{Iterative Co-Refinement}
The most recent frameworks establish a bi-directional dialogue where the LLM and statistical modules iteratively refine the structure.
Causal Modelling Agents (CMA) \citep{abdulaal2023causal} introduces an agentic loop where the LLM modifies the graph based on previous score history. This represents a shift towards dynamic collaboration. However, while CMA receives score feedback, its accuracy heavily depends on the agent's reasoning; if the LLM fails to predict a valid modification, there is no rigorous statistical constraint to prevent performance degradation. Furthermore, CMA places excessive reliance on the zero-shot initialization capabilities of LLMs. This strategy overlooks a critical reality: in many complex scenarios, initial skeletons inferred by established data-driven discovery models often exhibit higher fidelity to the observational data than purely semantic hypotheses generated by LLMs.

\section{Method}
\label{sec:method}

\begin{figure*}[htbp]
    \centering
    \includegraphics[width=1\linewidth]{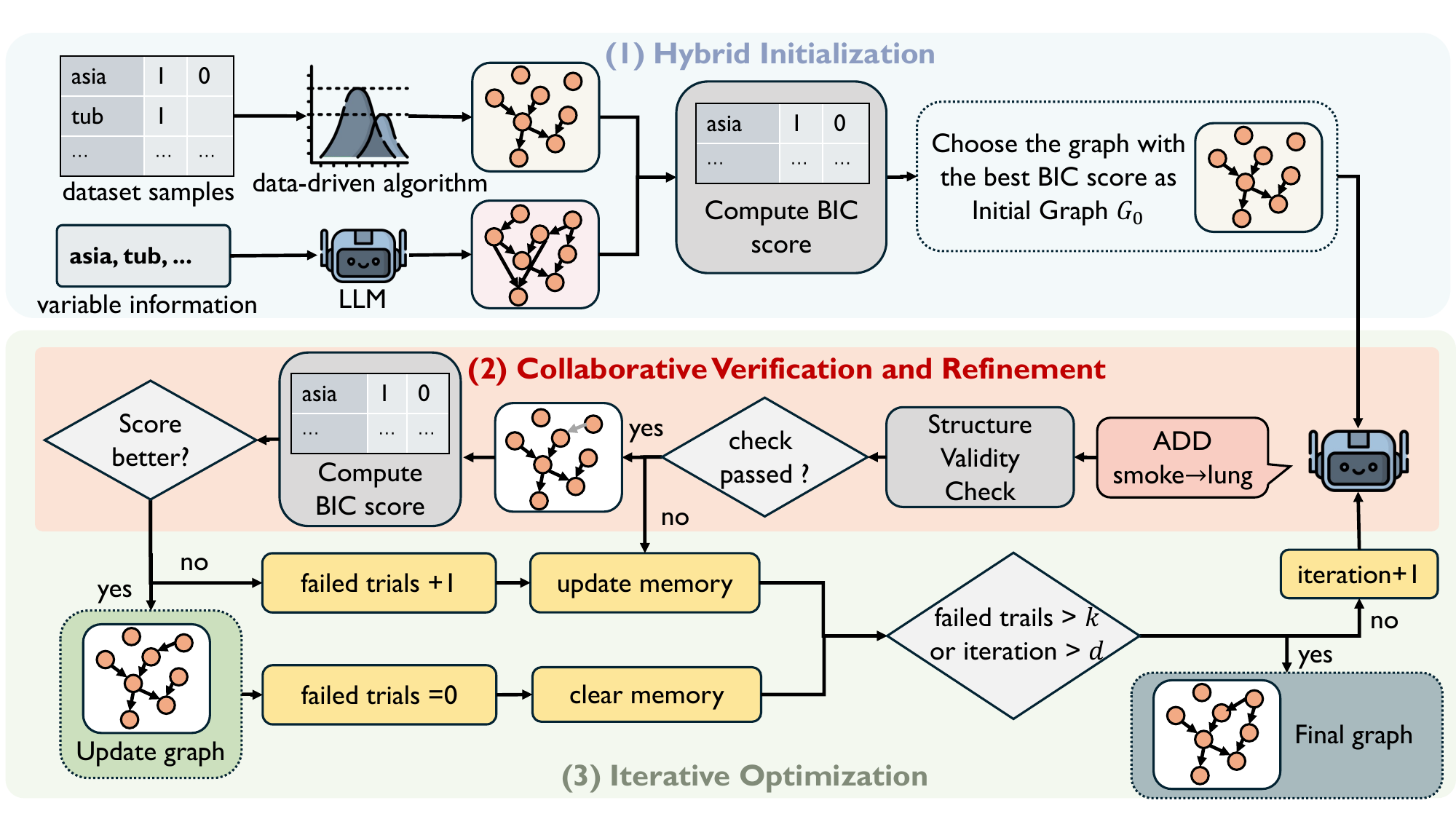} 
    \caption{Pipeline of \sys. The framework operates in three stages: \textbf{(1) Hybrid Initialization}, where the initial graph $\mathcal{G}_0$ is selected from either a data-driven baseline or an LLM hypothesis based on the superior BIC score; \textbf{(2) Collaborative Verification and Refinement}, where the LLM proposes atomic modifications (e.g., adding an edge) that are rigorously evaluated by a statistical verifier for structural validity and BIC improvement; and \textbf{(3) Iterative Optimization}, where valid proposals update the graph state while rejected ones populate an error memory to prevent the LLM from repeating invalid moves.}
    \label{fig:method}
\end{figure*}

The \sys framework comprises three components: (1) Hybrid Initialization for BIC-informed selection of the optimal starting graph (Section \ref{stage1}); (2) Collaborative verification and refinement, featuring a memory-augmented LLM-proposer and a statistical verifier to refine graph structures (Section \ref{stage2}); and (3) Iterative optimization for driving the causal discovery process toward final convergence (Section \ref{stage3}). 

\subsection{Intervention-Aware BIC}
\label{sec:global_bic}

The core of our \sys framework is the collaboration between the LLM and a rigorous statistical verifier. Formally, let 
% $\mathcal{D} = (\mathbf{X}_{obs}, \mathbf{X}_{int})$ 
$\mathcal{D} = (\mathcal{D}_{obs}, \mathcal{D}_{int})$
denote the dataset containing both observational and interventional samples over variables $\mathcal{V} = \{X_1, \dots, X_d\}$, where $d$ denotes the number of variables. Our goal is to uncover the causal graph $\mathcal{G}^*$ corresponds to the dataset.

To enable this verifier to provide objective, ground-truth feedback, we formulate a global fitness score based on the \textbf{Intervention-Aware Bayesian Information Criterion (BIC)}.
This scoring mechanism explicitly quantifies the trade-off mentioned in our objective: balancing \textit{data fidelity} (how well the graph explains the observed and interventional data) against \textit{structural complexity} (enforcing parsimony to prevent overfitting). Following the Minimum Description Length (MDL) principle~\citep{lam1994learning}, we define the score for a candidate graph $\mathcal{G}$ as:

\begin{equation*}
    \mathrm{BIC}(\mathcal{G}) = \underbrace{-2 \cdot \mathcal{L}_{\text{MLP}}(\mathcal{D} | \mathcal{G})}_{\text{Data Fidelity Term}} + \underbrace{k_{\text{eff}}(\mathcal{G}) \cdot \ln(N)}_{\text{Complexity Penalty Term}},
\end{equation*}
where $\mathcal{L}_{\text{MLP}}$ is the maximized log-likelihood estimated via neural networks, $k_{\text{eff}}$ is the effective parameter count, and $N$ is the sample size. We minimize this score to find the optimal structure.

\paragraph{Intervention-Aware Data Fidelity Score.} 
The first term measures how well the graph explains the data. We model the conditional probabilities using Multi-Layer Perceptrons (MLPs) to capture potential non-linear dependencies. 

Since our dataset $\mathcal{D}$ contains both observational and interventional samples, a standard likelihood calculation would be erroneous. A hard intervention on variable $X_i$ (denoted as $do(X_i=x)$) disrupts the natural causal mechanism, rendering the parent set $PA_i$ irrelevant for that specific sample. Including these samples in the evaluation would penalize the correct causal graph for failing to predict artificial intervention values.

To address this, we adopt the intervention-aware scoring principle from GIES~\citep{hauser2012characterization}, adapted here for non-linear mechanisms. We compute the log-likelihood only over samples where the causal mechanism is intact:
\begin{equation*}
    \mathcal{L}_{\text{MLP}}(\mathcal{D} | \mathcal{G}) = \sum_{i=1}^{d} \sum_{k=1}^{N} (1 - I_{k,i}) \cdot \log P_{\theta_i}(x_{k,i} | \mathbf{x}_{k, PA_i})
\end{equation*}
Here, $I_{k,i} \in \{0, 1\}$ is an indicator that equals 1 if variable $X_i$ is intervened in sample $k$, and 0 otherwise.  By zeroing out the contribution of intervened samples, this mask ensures the verifier is not penalized for failing to predict intervention artifacts, focusing solely on the validity of natural causal mechanisms. $P_{\theta_i}$ is modeled by a Multi-Layer Perceptron (MLP) to capture complex non-linear dependencies.

\paragraph{Structural Complexity ($k_{\text{eff}}$).} 
The second term imposes a sparsity constraint to prevent overfitting. We calculates the \textit{effective} parameter count $k_{\text{eff}}$ based on the theoretical degrees of freedom of the corresponding discrete Bayesian Network. For variables with cardinality $r$ (i.e., the number of unique discrete states), this is defined as:
\begin{equation*}
    k_{\text{eff}}(\mathcal{G}) = \sum_{i=1}^{d} (r_i - 1) \cdot \prod_{X_j \in PA_i} r_j
\end{equation*}
This formulation imposes a coherent statistical constraint: as the LLM adds edges, the penalty grows exponentially with the number of parents $PA$. This strong regularization forces the system to accept semantic proposals only when they provide a substantial gain in data fidelity, effectively filtering out weak or spurious associations suggested by the LLM.

\subsection{Hybrid Initialization}
\label{stage1}
To establish a robust starting point for the optimization process, we generate two candidate graphs leveraging distinct sources of information:
\begin{equation*}
    \mathcal{G}_{stat} \leftarrow \text{Baseline}(\mathcal{D})
\end{equation*}
\begin{equation*}
\mathcal{G}_{llm} \leftarrow \text{LLM}(\mathcal{V}_{\text{meta}})
\end{equation*}

\noindent (i) $\mathcal{G}_{stat}$: 
Derived from a standard data-driven baseline (e.g., FCI~\citep{10.5555/2074158.2074215} or AVICI~\citep{lorch2022amortized}). This candidate captures statistical dependencies but may struggle with overly conservative analysis or errors caused by domain shift.

\noindent (ii) $\mathcal{G}_{llm}$: 
Generated by the LLM based on variable information (e.g., variable names). This candidate provides a semantic prior but may contain hallucinations or domain misalignments.

We employ the previously defined Intervention-Aware BIC as an automated criterion to evaluate both candidates. The candidate yielding the lower score is instantiated as the initial structure $\mathcal{G}_0$:
\begin{equation*}
    \mathcal{G}_0 \leftarrow \mathop{\arg\min}_{\mathcal{G} \in \{\mathcal{G}_{stat}, \mathcal{G}_{llm}\}} \mathrm{BIC}(\mathcal{G}, \mathcal{D})
\end{equation*}

By dynamically selecting the superior candidate, this hybrid strategy \textbf{bypasses the inherent limitations of relying on a single modality}: it avoids initializing with a severely flawed structure from a pure data-based method or a hallucinated guess from an LLM. This ensures that the subsequent iterative refinement starts from a relatively reliable initialization.

\subsection{Collaborative Verification and Refinement}
\label{stage2}
In this stage, \sys iteratively refines the graph through a collaborative loop between the LLM and a statistical verifier. The LLM modifies the graph structure via a set of atomic operations $\mathcal{A} = \{ \text{ADD}(i, j), \text{DEL}(i, j), \text{REV}(i, j) \}$, representing the addition, deletion, or reversal of a directed edge $X_i \to X_j$, respectively.

\paragraph{Verification.}
Upon receiving a proposed modification $a_{t}$, the verifier evaluates it in two steps.

\noindent (i) Structural validity check.
The proposal is \textbf{immediately rejected} if it is not a valid transformation, including:
(1) introducing a directed cycle;
(2) deleting or reversing an edge that does not exist in $\mathcal{G}_t$;
(3) adding an edge that already exists in $\mathcal{G}_t$; or
(4) referencing variables outside the dataset domain.

\noindent (ii) Statistical improvement check.
If structurally valid, we apply the edit to obtain $\mathcal{G}' = a_t(\mathcal{G}_t)$ and compute its intervention-aware BIC score.
We accept the edit only if it improves the score:
\begin{equation*}
\Delta \mathrm{BIC} = \mathrm{BIC}(\mathcal{G}_t) - \mathrm{BIC}(\mathcal{G}') > 0,
\end{equation*}
\noindent\textit{(Note: Since we aim to minimize BIC, a positive reduction $\Delta \mathrm{BIC} > 0$ indicates improvement.)}

\paragraph{Refinement.}
To facilitate efficient collaboration, we maintain an error memory $\mathcal{M}_{\text{err}}$ that logs rejected edits, including the operation type and the rejection reason (structural violation or $\Delta \mathrm{BIC}\le 0$).
At each iteration, the LLM is prompted with (i) the current graph state (variable list and edges in $\mathcal{G}_t$) and (ii) $\mathcal{M}_{\text{err}}$, which acts as a lightweight negative constraint to prune the search space and avoid repeating invalid or unhelpful edits.

\subsection{Iterative Optimization}
\label{stage3}

The optimization maintains an error memory $\mathcal{M}_{\mathrm{err}}$ and a counter $c$ for consecutive \emph{statistical} rejections (used for early stopping). At each iteration:

\noindent (i) Reject. 
If $a_t$ fails the structural check or $\Delta \mathrm{BIC}\le 0$, we keep $\mathcal{G}_{t+1}\leftarrow \mathcal{G}_t$ and log the failure:
    \begin{equation*}
        \mathcal{M}_{\mathrm{err}} \leftarrow \mathcal{M}_{\mathrm{err}} \cup \{(a_t,\mathrm{Reason}_t)\}.
    \end{equation*}
We increment $c\leftarrow c+1$ only when the rejection is statistical (i.e., $\Delta \mathrm{BIC}\le 0$); otherwise $c$ is unchanged.

\noindent (ii) Accept.
If $\Delta \mathrm{BIC}>0$, we update $\mathcal{G}_{t+1}\leftarrow \mathcal{G}'$, reset $c\leftarrow 0$, and clear the memory $\mathcal{M}_{\mathrm{err}}\leftarrow \emptyset$ to avoid carrying stale constraints after the structure changes.
We iterate for at most $T$ steps (set to $T=d$ in our experiments) and stop early when $c\ge k$ (we use $k=5$). Finally, we return the graph with the lowest BIC encountered during the search.

\section{Experiments}
\label{sec:experiments}

\subsection{Experimental Setup}

\paragraph{Datasets}
We evaluate our framework on four standard benchmark datasets from the Bayesian Network Repository~\cite{bnrepository}: \textit{Cancer}~\cite{KorbNicholson2010} (5 nodes), \textit{Asia}~\cite{10.1111/j.2517-6161.1988.tb01721.x} (8 nodes), \textit{Child}~\cite{10.1093/oso/9780198522669.003.0025} (20 nodes), and \textit{Alarm}~\cite{10.1007/978-3-642-93437-7_28} (37 nodes). These datasets cover a range of complexities, from small networks to medium-scale networks with varying edge densities. For each dataset, we utilize a mix of observational and interventional samples to simulate a realistic causal discovery scenario. We generate discrete data and intervene on each nodes one at a time, we sample 5000 data points for each dataset.

\paragraph{Baselines}
We compare \sys against two categories of methods: 

\begin{itemize}
    \item \textbf{Data-driven Algorithms:} \textit{FCI} (Fast Causal Inference)~\cite{10.5555/2074158.2074215}, a widely used constraint-based method, and \textit{AVICI}~\cite{lorch2022amortized}, a recent amortization-based causal discovery model.
    \item \textbf{Zero-shot LLMs:} \textit{Qwen3-14B} and \textit{Qwen3-32B}~\cite{qwen3technicalreport}, prompted to generate the causal graph directly from variable information without statistical verification.
\end{itemize}

\paragraph{Metrics}
We report standard causal discovery metrics: Precision, Recall, F1-Score, and Structural Hamming Distance (SHD).

\begin{table*}[t]
\centering
% =======================================================
% 宏定义：只显示均值 (#1)
% =======================================================
\newcommand{\res}[2]
{#1} 
\newcommand{\gres}[2]
{\setlength{\fboxsep}{1.5pt}\colorbox{morandiorange}{#1}}
% =======================================================
\newcommand{\stats}[2]{($d{=}#1, |E|{=}#2$)}
\setlength{\tabcolsep}{2.5pt}

% =======================================================
% Part 1: Cancer & Asia
% =======================================================
\resizebox{\textwidth}{!}{%
\begin{tabular}{lcccccccc}
\toprule
\multirow{2}{*}{\textbf{Method}} & \multicolumn{4}{c}{\textbf{Cancer} \stats{5}{4}} & \multicolumn{4}{c}{\textbf{Asia} \stats{8}{8}} \\
\cmidrule(lr){2-5} \cmidrule(lr){6-9} 
& \textbf{Precision} ($\uparrow$) & \textbf{Recall} ($\uparrow$) & \textbf{F1-Score} ($\uparrow$) & \textbf{SHD} ($\downarrow$) & \textbf{Precision} ($\uparrow$) & \textbf{Recall} ($\uparrow$) & \textbf{F1-Score} ($\uparrow$) & \textbf{SHD} ($\downarrow$) \\
\midrule
\multicolumn{9}{l}{\textit{\textbf{Reference: Pure LLM (Zero-shot)}}} \\
Qwen3-14B & \res{\textbf{100.0}}{0} & \res{\textbf{100.0}}{0} & \res{\textbf{100.0}}{0} & \res{\textbf{0.0}}{0} & \res{67.4}{19.0} & \res{82.5}{10.0} & \res{73.7}{15.3} & \res{5.0}{3.0} \\
Qwen3-32B & \res{\textbf{100.0}}{0} & \res{\textbf{100.0}}{0} & \res{\textbf{100.0}}{0} & \res{\textbf{0.0}}{0} & \res{90.6}{8.5} & \res{92.5}{6.1} & \res{91.5}{7.2} & \res{1.4}{1.2} \\
\midrule
\multicolumn{9}{l}{\textit{\textbf{FCI-based Methods}}} \\
FCI (Baseline) & 20.0 & \textbf{100.0} & 33.3 & 16.0 & 66.7 & 25.0 & 36.4 & 6.0 \\
\textbf{+ Ours} (Qwen3-14B) & \gres{63.3}{6.7} & \gres{\textbf{100.0}}{0} & \gres{77.3}{5.3} & \gres{2.4}{0.8} & \gres{76.5}{8.2} & \gres{87.5}{11.2} & \gres{81.3}{8.5} & \gres{3.2}{1.5} \\
\textbf{+ Ours} (Qwen3-32B) & \gres{77.3}{5.3} & \gres{\textbf{100.0}}{0} & \gres{87.1}{3.6} & \gres{1.2}{0.4} & \gres{83.1}{13.5} & \gres{92.5}{10.0} & \gres{87.5}{11.9} & \gres{2.2}{2.1} \\
\midrule
\multicolumn{9}{l}{\textit{\textbf{AVICI-based Methods}}} \\
AVICI (Baseline) & \res{\textbf{100.0}}{0} & \res{35.0}{12.2} & \res{50.7}{13.1} & \res{2.6}{0.5} & \res{\textbf{100.0}}{0} & \res{57.5}{6.1} & \res{72.8}{5.0} & \res{3.4}{0.5} \\
\textbf{+ Ours} (Qwen3-14B) & \gres{\textbf{100.0}}{0} & \gres{\textbf{100.0}}{0} & \gres{\textbf{100.0}}{0} & \gres{\textbf{0.0}}{0} & \res{97.8}{4.4} & \gres{92.5}{10.0} & \gres{94.6}{5.3} & \gres{0.8}{0.8} \\
\textbf{+ Ours} (Qwen3-32B) & \gres{\textbf{100.0}}{0} & \gres{\textbf{100.0}}{0} & \gres{\textbf{100.0}}{0} & \gres{\textbf{0.0}}{0} & \gres{\textbf{100.0}}{0} & \gres{\textbf{95.0}}{6.1} & \gres{\textbf{97.3}}{3.3} & \gres{\textbf{0.4}}{0.5} \\
\bottomrule
\end{tabular}%
}
\vspace{10pt}
% =======================================================
% Part 2: Child & Alarm
% =======================================================
\resizebox{\textwidth}{!}{%
\begin{tabular}{lcccccccc}
\toprule
\multirow{2}{*}{\textbf{Method}} & \multicolumn{4}{c}{\textbf{Child} \stats{20}{25}} & \multicolumn{4}{c}{\textbf{Alarm} \stats{37}{46}} \\
\cmidrule(lr){2-5} \cmidrule(lr){6-9} 
& \textbf{Precision} ($\uparrow$) & \textbf{Recall} ($\uparrow$) & \textbf{F1-Score} ($\uparrow$) & \textbf{SHD} ($\downarrow$) & \textbf{Precision} ($\uparrow$) & \textbf{Recall} ($\uparrow$) & \textbf{F1-Score} ($\uparrow$) & \textbf{SHD} ($\downarrow$) \\
\midrule
\multicolumn{9}{l}{\textit{\textbf{Reference: Pure LLM (Zero-shot)}}} \\
Qwen3-14B & \res{49.3}{6.8} & \res{46.4}{5.9} & \res{47.6}{5.6} & \res{23.6}{3.1} & \res{38.5}{3.1} & \res{35.2}{8.6} & \res{36.4}{5.6} & \res{54.2}{2.3} \\
Qwen3-32B & \res{49.7}{5.9} & \res{\textbf{55.2}}{8.5} & \res{51.5}{2.7} & \res{25.4}{2.3} & \res{33.0}{2.5} & \res{35.2}{3.5} & \res{33.9}{1.3} & \res{61.8}{5.2} \\
\midrule
\multicolumn{9}{l}{\textit{\textbf{FCI-based Methods}}} \\
FCI (Baseline) & 37.5 & 12.0 & 18.2 & 22.0 & \textbf{100.0} & 34.8 & 51.6 & 30.0 \\
\textbf{+ Ours} (Qwen3-14B) & \gres{40.6}{4.3} & \gres{20.0}{6.2} & \gres{26.5}{5.8} & \gres{21.6}{1.4} & \res{84.8}{3.8} & \gres{47.4}{4.2} & \gres{60.6}{3.1} & \gres{27.4}{1.6} \\
\textbf{+ Ours} (Qwen3-32B) & \gres{44.9}{7.7} & \gres{17.6}{2.0} & \gres{25.2}{3.0} & \gres{21.8}{1.5} & \res{76.7}{6.6} & \gres{45.7}{6.2} & \gres{56.7}{4.2} & \res{31.2}{2.0} \\
\midrule
\multicolumn{9}{l}{\textit{\textbf{AVICI-based Methods}}} \\
AVICI (Baseline) & \res{\textbf{100.0}}{0} & \res{24.0}{5.1} & \res{38.4}{6.6} & \res{19.0}{1.3} & \res{95.2}{2.2} & \res{58.7}{3.0} & \res{72.5}{2.0} & \res{19.4}{1.0} \\
\textbf{+ Ours} (Qwen3-14B) & \res{67.7}{13.3} & \gres{52.8}{11.1} & \gres{\textbf{57.4}}{7.6} & \gres{\textbf{18.6}}{2.9} & \gres{96.5}{0.1} & \gres{59.6}{1.7} & \gres{73.6}{1.4} & \gres{18.6}{0.8} \\
\textbf{+ Ours} (Qwen3-32B) & \res{73.9}{21.0} & \gres{45.6}{10.3} & \gres{53.6}{6.7} & \res{19.0}{4.7} & \gres{96.7}{2.4} & \gres{\textbf{63.0}}{1.5} & \gres{\textbf{76.3}}{1.6} & \gres{\textbf{17.8}}{0.8} \\
\bottomrule
\end{tabular}%
}
\caption{Performance comparison with full metrics. All experiments were repeated 5 times, and the average performance is reported. Datasets are annotated with their complexity (number of nodes $d$ and number of edges $|E|$). The table is split into two panels: \textbf{(Top)} small-scale networks (Cancer, Asia); \textbf{(Bottom)} medium-scale networks (Child, Alarm). Arrows indicate the direction of better performance ($\uparrow$: Higher is better, $\downarrow$: Lower is better). \textbf{Bold} indicates the best result. The \colorbox{morandiorange}{highlight} indicates our method improves the pure-data baseline.}
\label{tab:model_comparison_full_stats}
\end{table*}

\subsection{Main Results}
\label{sec:main_results}
Table~\ref{tab:model_comparison_full_stats} summarizes the quantitative performance of our proposed \sys framework against current baselines. The results demonstrate the effectiveness of our method across varying graph complexities.

\paragraph{Universal Enhancement on Pure Data-based Algorithms.}
The most prominent finding is that \textbf{\sys consistently improves performance regardless of the underlying pure-data causal discovery baseline} (\textit{FCI} or \textit{AVICI}), acting as a universal ``enhancer.'' Our method achieves the lowest SHD across nearly all configurations. For instance, on the \textit{Cancer} dataset, integrating \sys with FCI reduces the SHD from 16 to 1.2 (using Qwen3-32B). Similarly, on the \textit{Asia} dataset, it reduces the SHD of the FCI baseline from 6 to 2.2. Furthermore, the framework consistently boosts the F1-score; notably, on the \textit{Cancer} dataset, the FCI baseline's F1-score improves from 33.3 to 87.1 with the addition of our method.

\paragraph{Addressing the Unreliability of LLMs in Complex Tasks.}
The results highlight the limitations of direct LLM inference and how \sys resolves them. \textbf{While Zero-shot models perform well on simple datasets, this success is highly fragile, giving way to unacceptably low accuracy on even moderately complex tasks.} On the \textit{Alarm} network (37 nodes), pure Qwen3-32B yields a high SHD of 61.8. \sys drastically mitigates these errors. For the \textit{Alarm} dataset, the \textit{AVICI + Ours (Qwen3-32B)} configuration achieves a much lower SHD of 17.8, proving that introducing pure-data baseline and BIC verification strategy effectively reduce the errors in LLMs.

\begin{figure*}[t]
    \centering
    \includegraphics[width=1\linewidth]{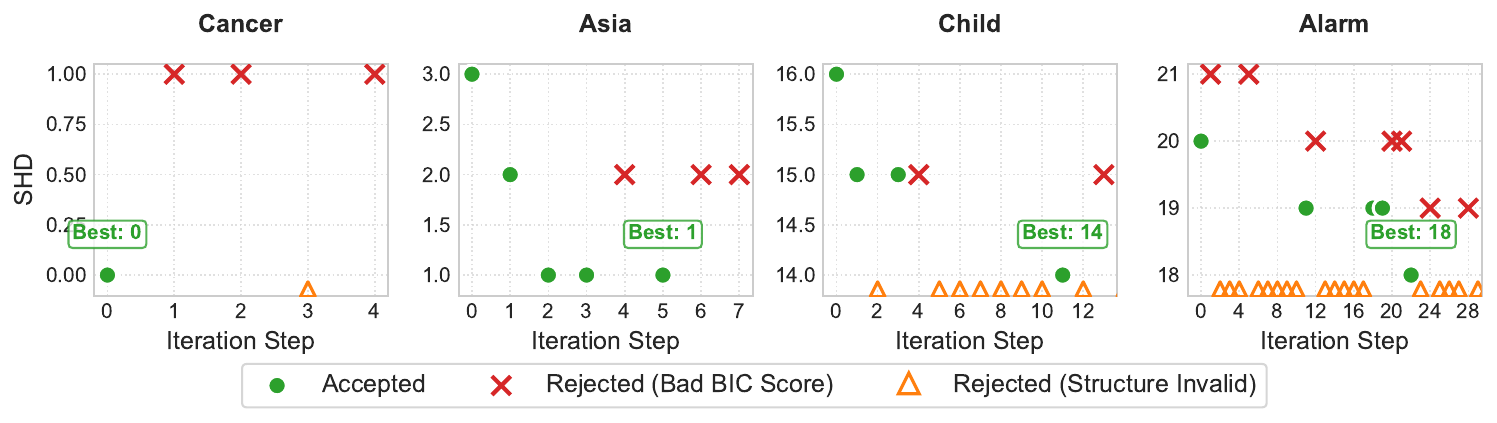} 
    \caption{Optimization Trajectories of Qwen3-14B with AVICI as data-driven algorithm. LLM proposes reasonable edges operations during optimization loop (green circles), while BIC varifier successfully rejected operations with statistical inconsistency (red crosses). Note that for structure invalid errors (orange triangles), SHD is not computed. Therefore, we illustrate these marks on the x-axis.}
    \label{fig:optimization_trajectory}
\end{figure*}

\paragraph{Complementing Statistical Limitations.}
A key insight is \sys's ability to compensate for the specific weaknesses of learning-based methods. The \textit{AVICI Baseline} tends to be conservative, achieving high precision (100.0 on Cancer/Asia/Child) but suffering from low recall (e.g., 35.0 on Cancer, 24.0 on Child). \textbf{\sys identifies semantic edges that pure statistics miss, significantly improving the Recall.} When combined with AVICI, \sys boosts Recall on \textit{Cancer} from 35.0 to 100.0 and on \textit{Asia} from 57.5 to 95.0 (Qwen3-32B) while maintaining high Precision.

\paragraph{Robustness on Larger Graphs.}
The results on the \textit{Child} and \textit{Alarm} datasets demonstrate the framework's scalability. The \textit{AVICI + Ours (Qwen3-32B)} configuration achieves the best performance on the most complex dataset, \textit{Alarm}, with the highest F1-score (76.3) and lowest SHD (17.8), surpassing both the pure statistical baseline (SHD 19.4) and the pure LLM (SHD 61.8). Even on the challenging \textit{Child} dataset, \sys improves the F1-score of AVICI from 38.4 to 57.4 (Qwen3-14B), validating the effectiveness of the iterative propose-and-verify mechanism in larger search spaces.

\subsection{Optimization Trajectories}
\label{sec:optimization_dynamics}

To distinguish the roles of the LLM agent and the verification module, we analyze the optimization trajectories of Qwen3-14B with AVICI as data-driven algorithm in Figure~\ref{fig:optimization_trajectory}.

\paragraph{LLM Act as Hypothesis Generator.}
Taking the \textit{Child} dataset as example, the result highlights the framework's ability to navigate a larger search space where baseline methods often miss subtle dependencies. 
Starting with a high structural error (SHD=16), the agent leverages semantic knowledge to propose missing links that the initial pure data-based baseline failed to capture. 
The trajectory shows a monotonic improvement, reducing the SHD to 14. 
This phase demonstrates the \textbf{discovery potential} of the system: the LLM acts as a reasoning engine to break out of the local optima trapped by the traditional algorithm.

\paragraph{BIC Score as Reliable Verifier.}
We further illustrates the necessity of a strict Statistical Veto. For example, in \textit{Asia}, after the model rapidly converges to a near-perfect structure (SHD=1), the LLM continues to propose operations on edges based on plausible but statistically unsupported associations. In this case, our \textbf{BIC-based Verifier successfully identified and rejected these false operations.} 
Unlike soft-feedback mechanisms that might succumb to cumulative errors, our method enforces a hard stop, prioritizing data fidelity over the LLM's generative tendencies.

\subsection{Score Function Validity}
\label{sec:score_validity}

\begin{figure}[t]
    \centering
    \includegraphics[width=1\linewidth]{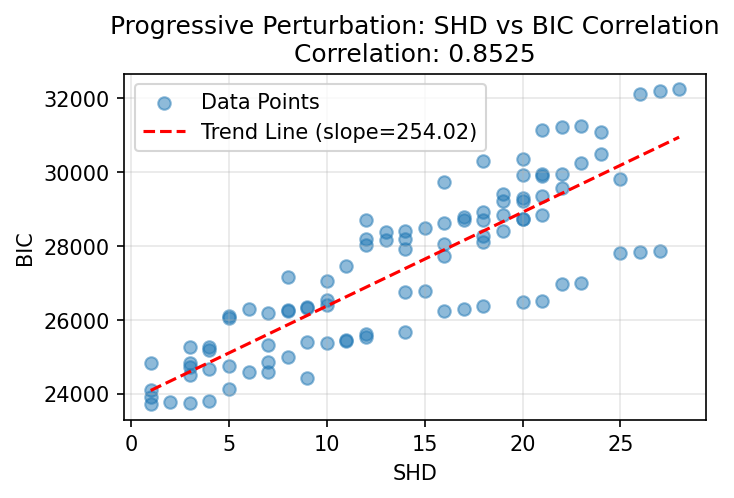}
    \caption{Score Validity on Alarm Dataset. We plot 100 perturbed graphs generated via 5 random walk trajectories. The X-axis represents structural error (SHD), and the Y-axis represents the intervention-aware BIC. The strong positive trend confirms that our scoring function effectively penalizes structural errors.}
    \label{fig:perturbation_alarm}
\end{figure}

The core premise of our \sys framework is that the intervention-aware BIC serves as a reliable proxy for structural fidelity. 
To ensure its validity, we conducted a progressive perturbation experiment on the \textit{Alarm} dataset. 
Starting from the ground truth graph $\mathcal{G}^*$, we generated 5 independent random walk trajectories. 
In each trajectory, we performed 20 steps of cumulative modifications, where each step involved a random atomic operation (edge addition, removal, or reversal) applied to the previous state. In total, 100 perturbation graphs are generated.
We then recorded the SHD and compute the BIC score in each step.
As illustrated in Figure~\ref{fig:perturbation_alarm}, \textbf{we observe a significant positive correlation ($\rho \approx$ 0.852) between SHD and BIC, which validates BIC as a reliable signal for assessing LLM hypotheses.}

\subsection{LLM Hypothesis Analysis}
\label{sec:error_analysis}

\begin{figure}[t]
    \centering
    \includegraphics[width=1\linewidth]{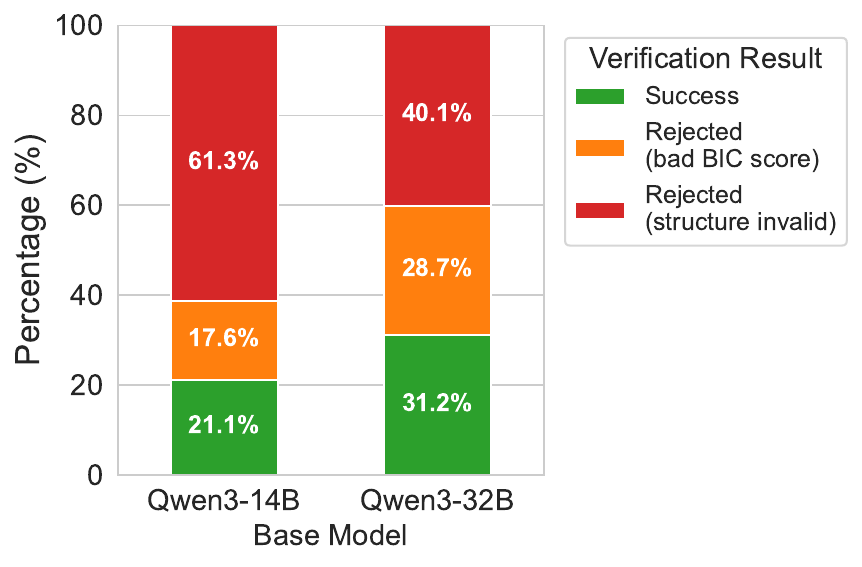} 
    \caption{Analysis of LLM optimization trajectories categorized by verification outcome. Larger models (e.g., Qwen3-32B) demonstrate superior reasoning capabilities, resulting in fewer structural violations and higher acceptance rates.}
    \label{fig:llm_error}
\end{figure}

We conducted a fine-grained analysis of the optimization trajectory by aggregating all atomic operations across datasets and categorizing them into three outcomes: \textit{Success}, \textit{Rejected (structure invalid)}, and \textit{Rejected (bad BIC score)} (worsened BIC score). As shown in Figure~\ref{fig:llm_error}, \textit{Structure Invalid} errors constitute the largest portion of failures (61.3\% for Qwen3-14B), empirically validating the necessity of our explicit structural validity check. Furthermore, the larger model (Qwen3-32B) demonstrates significantly better structure validity awareness, reducing the structure invalid rate from 61.3\% to 40.1\%. Furthermore, scaling the model from 14B to 32B boosts the success rate from 21.1\% to 31.2\%, demonstrating that larger models possess superior reasoning capabilities for generating high-quality hypotheses under constraints.

\section{Conclusion}
In this work, we addressed the fundamental challenge of integrating semantic knowledge with statistical rigor for causal discovery. We proposed \sys, a collaborative framework that synergizes LLMs as hypothesis-generating ``data scientists'' with probabilistic statistics as rigorous ``verifiers''. Through hybrid initialization, collaborative verification and refinement, and iterative optimization, \sys bridges the gap between rich causal knowledge encoded in LLMs and the empirical constraints of observational data.

\section*{Limitations}
First, our method relies on the semantic reasoning of LLMs, making it most effective in domains with rich descriptive metadata. In scenarios where variables are anonymized (e.g., node names are masked), the LLM cannot leverage domain knowledge to generate informative priors. Although our statistical verification mechanism prevents LLM from introducing errors, the performance gain from the semantic component would naturally be limited in such knowledge-scarce environments.

Second, our current verification mechanism depends exclusively on the BIC score. While effective for penalizing complexity, BIC is an asymptotic criterion that may not be optimal for all sample sizes or data distributions. However, a key strength of our method is its flexibility: the statistical verification module can be easily substituted with other scoring objectives. Future implementations could incorporate alternative scoring functions to adapt to broader data regimes.
% Bibliography entries for the entire Anthology, followed by custom entries
%\bibliography{anthology,custom}
% Custom bibliography entries only
\bibliography{custom}
\clearpage
\appendix
\section{Implementation Details}
\label{appendix:implementation_details}

In this section, we provide the technical specifications and hyperparameter configurations used in our experiments to ensure reproducibility. We set iterations to $d$, which is the number of nodes for each dataset. We set the early stopping patience to 5 to allow for retries that cause worser BIC score.

\subsection{LLM Configuration}
For all runs, the decoding temperatures for Qwen3-14B and Qwen3-32B were set to $0.6$ to balance creativity and logical consistency. We utilized the \texttt{vLLM} inference engine with \texttt{Flash Attention} enabled to accelerate the generation process of the LLM. 

\subsection{Baseline Implementation}
\label{subsec:baseline_impl}

To incorporate data-driven insights, we integrated two distinct types of baseline causal discovery methods: \textit{AVICI} and \textit{FCI}. These baselines provide initial structural priors that guide the LLM's hypothesis generation.

\paragraph{AVICI} 
We utilize the \textit{AVICI} \cite{lorch2022amortized}, which is a deep learning-based approach. Our implementation uses the official \texttt{avici} repository with the \texttt{scm-v0} pretrained checkpoint. This model employs a Transformer-based architecture trained to predict causal structures directly from observational and interventional data matrices in a single forward pass. For the baseline method AVICI, we extract all predicted edges whose confidence scores exceed a predefined threshold of $0.5$. The resulting structure is then post-processed to remove any cycles, ensuring a Directed Acyclic Graph (DAG) is provided as a reference to the LLM.

\paragraph{FCI}
The \textit{FCI} algorithm is implemented via the \texttt{causal-learn} library. As a constraint-based method, FCI identifies causal relationships by performing conditional independence tests. Specifically, we use the \textbf{Chi-square test} for discrete data, with a significance level of $\alpha = 0.05$. Although FCI outputs a Partial Ancestral Graph (PAG) that may contain various edge types (such as undirected edges or edges with ambiguous endpoints), we adopt a conservative selection policy: only definitive directed edges ($i \to j$) are extracted and provided as prior knowledge to the LLM. All ambiguous or non-directed relationships are intentionally excluded to ensure the precision of the initial graph priors. Finally, any remaining cycles are resolved to maintain DAG consistency.

\subsection{Cycle Detection and Removal Algorithms}
\label{subsec:cycle}
\paragraph{Cycle Detection (DFS)}
We utilize a Depth-First Search (DFS) traversal to verify the acyclic nature of the graph. As detailed in Algorithm \ref{alg:detect_cycle}, the algorithm maintains a recursion stack to track the active traversal path. If the search encounters a node that is already present in the current recursion stack, a \textbf{back-edge} is identified. This confirms the existence of a cycle, and the algorithm immediately captures the specific sequence of nodes constituting the loop.

\paragraph{Cycle Breaking Strategy}
Upon detecting a cycle, arbitrarily removing an edge could disrupt significant causal relationships. To mitigate this, we employ a \textbf{greedy minimization strategy} based on edge weights. As described in Algorithm \ref{alg:break_cycle}, the procedure iterates strictly through the edges that form the detected cycle path. It identifies the ``weakest link''---the edge with the minimum weight---and removes it. This process is repeated iteratively until the graph is fully converted into a DAG, ensuring that the strongest causal signals are preserved.

\begin{algorithm}[h]
\caption{Cycle Detection via Depth-First Search}
\label{alg:detect_cycle}
\begin{algorithmic}[1]
\Require Directed Graph $G = (V, E)$
\Ensure Returns $(True, Path)$ if a cycle exists, else $(False, \emptyset)$
\State $Visited \gets \emptyset$
\State $RecursionStack \gets \emptyset$

\Function{DFS}{$u, path$}
    \State $Visited \gets Visited \cup \{u\}$
    \State $RecursionStack \gets RecursionStack \cup \{u\}$
    \State Append $u$ to $path$
    
    \For{each neighbor $v$ of $u$}
        \If{$v \notin Visited$}
            \If{\Call{DFS}{$v, path$} is \textbf{True}}
                \State \Return \textbf{True}
            \EndIf
        \ElsIf{$v \in RecursionStack$}
            \Comment{Back-edge detected: Cycle found}
            \State $start \gets$ index of $v$ in $path$
            \State $CyclePath \gets path[start:] + [v]$
            \State \Return \textbf{True}
        \EndIf
    \EndFor
    
    \State $RecursionStack \gets RecursionStack \setminus \{u\}$
    \State Remove last element from $path$
    \State \Return \textbf{False}
\EndFunction
\end{algorithmic}
\end{algorithm}

\begin{algorithm}[h]
\caption{Iterative Cycle Breaking (Min-Weight Strategy)}
\label{alg:break_cycle}
\begin{algorithmic}[1]
\Require Graph $G$, Edge Weights $W$
\Ensure A DAG where all cycles are resolved
\State $(detected, cycle\_path) \gets \Call{DetectCycle}{G}$

\While{$detected$ is \textbf{True}}
    \State $min\_w \gets \infty$
    \State $target\_edge \gets \textbf{null}$
    
    \Comment{Iterate only through edges belonging to the cycle}
    \For{$i \gets 0$ to length($cycle\_path$) $- 2$}
        \State $u \gets cycle\_path[i]$
        \State $v \gets cycle\_path[i+1]$
        \If{$W(u, v) < min\_w$}
            \State $min\_w \gets W(u, v)$
            \State $target\_edge \gets (u, v)$
        \EndIf
    \EndFor
    
    \If{$target\_edge \neq \textbf{null}$}
        \State Remove $target\_edge$ from $G$
    \EndIf
    
    \State $(detected, cycle\_path) \gets \Call{DetectCycle}{G}$
\EndWhile
\State \Return $G$
\end{algorithmic}
\end{algorithm}

\subsection{Evaluation Metrics}
The quality of the discovered causal graphs was evaluated using several standard metrics:
\begin{itemize}
    \item \textbf{SHD (Structural Hamming Distance):} Measures the number of edge additions, deletions, and reversals required to transform the predicted graph into the ground truth.
    \item \textbf{Precision, Recall, and F1-score:} Evaluated based on the existence and direction of the predicted edges.
\end{itemize}

\section{BIC Score}
\label{app:bic_implementation}

To robustly evaluate the quality of candidate causal graphs $\mathcal{G}$ on discrete data, we employ a hybrid scoring mechanism. We utilize a neural network to estimate the likelihood of the data given the graph structure, while calculating the complexity penalty term based on the theoretical degrees of freedom of a discrete Bayesian Network. This approach combines the universal approximation capabilities of Multi-Layer Perceptrons (MLPs) with the statistical rigor of the Bayesian Information Criterion (BIC).

\subsection{Neural Likelihood Estimation}
We adopt the \texttt{MultivarMLP} architecture adapted from the ENCO framework~\citep{lippe2022enco}. For a dataset $\mathcal{D} = \{ \mathbf{x}^{(k)} \}_{k=1}^{N}$ with $d$ discrete variables, we model the conditional probability distributions $P(X_i | PA_i)$ using a shared embedding layer followed by parallel MLPs, where $PA_i$ denotes the set of parents of variable $X_i$ in graph $\mathcal{G}$.

\paragraph{Masking and Forward Pass} 
The structural constraints of $\mathcal{G}$ are enforced via an adjacency mask. The input to the MLP for variable $X_i$ is masked such that it only receives information from $PA_i$. For discrete variables, we map categorical indices to continuous dense vectors using a learnable embedding matrix. The network outputs the logits for the categorical distribution of each variable.

\paragraph{Optimization}
We train the parameters $\theta$ of the MLP to minimize the negative log-likelihood (NLL). For discrete data, this is equivalent to the Cross-Entropy loss:
\begin{equation*}
    \mathcal{L}_{\text{MLP}}(\mathcal{D} | \mathcal{G}) = - \sum_{k=1}^{N} \sum_{i=1}^{d} (1 - I_{k,i}) \cdot \log P_{\theta}(x_{k,i} | \mathbf{x}_{k, PA_i})
\end{equation*}
where $I_{k,i}$ is an indicator function that equals 1 if variable $X_i$ was intervened upon in sample $k$, and 0 otherwise. This ensures that the score reflects the fit of the causal mechanisms rather than the intervention policy. We typically train the MLP for 100 epochs to ensure the convergence of probability estimates.

\subsection{Effective Parameter Counting ($k_{\text{eff}}$)}
A critical component of our implementation is the calculation of the complexity penalty. Using the raw number of neural network weights would result in severe over-penalization. Instead, we calculate the \textit{effective number of parameters} $k_{\text{eff}}$ corresponding to a discrete Bayesian Network with the structure $\mathcal{G}$.

For each variable $X_i$ with cardinality $r_i$ (number of unique states), and a parent set $PA_i$ where each parent $X_j$ has cardinality $r_j$, the number of independent parameters required to specify the Conditional Probability Table (CPT) is:
\begin{equation*}
    k_i = (r_i - 1) \cdot \prod_{X_j \in PA_i} r_j
\end{equation*}
The total effective degrees of freedom for the graph is the sum over all nodes: $k_{\text{eff}} = \sum_{i=1}^{d} k_i$. This count accurately reflects the statistical complexity of the graph structure.

\subsection{Final BIC Score}
The final score for a candidate graph $\mathcal{G}$, which we aim to minimize, is defined as:
\begin{equation*}
    \text{BIC}(\mathcal{G}) = -2 \cdot \hat{\mathcal{L}}_{\text{MLP}}(\mathcal{D} | \mathcal{G}) + k_{\text{eff}} \cdot \ln(N)
\end{equation*}
where $\hat{\mathcal{L}}_{\text{MLP}}$ is the maximized log-likelihood estimated by the neural network, and $N$ is the sample size. This scoring function allows us to leverage the flexibility of neural networks to capture complex dependencies while maintaining a valid statistical penalty to prevent overfitting.

\subsection{Hyperparameter Configuration}
\label{app:hyperparams}

For reproducibility, we detail the specific architecture and optimization hyperparameters used for the Neural BIC scoring model in Table \ref{tab:mlp_hyperparams}. 

The conditional probability distributions are modeled using a Multi-Layer Perceptron (MLP). We map discrete variables to a dense vector space of dimension 64. The network consists of one hidden layer with 64 units, employing LeakyReLU activation ($\text{negative slope}=0.1$) to prevent dying gradients. Weights are initialized using Kaiming Uniform initialization. The model is trained using the Adam optimizer with a fixed learning rate of $1e-2$ for 100 epochs, which we found sufficient for convergence on all benchmark datasets.

\begin{table}[h]
    \centering
    \caption{Hyperparameters for the MLP-based BIC Scoring Model.}
    \label{tab:mlp_hyperparams}
    \begin{tabular}{l|c}
    \hline
    \textbf{Hyperparameter} & \textbf{Value} \\
    \hline
    Embedding Dimension & 64 \\
    Hidden Layer Size & [64] \\
    Activation Function & LeakyReLU ($\alpha=0.1$) \\
    Optimizer & Adam \\
    Learning Rate & 0.01 \\
    Training Epochs & 100 \\
    Weight Initialization & Kaiming Uniform \\
    \hline
    \end{tabular}
\end{table}

\section{Prompt}
We illustrate our prompt below. In our experiments, the zero-shot user prompt and graph refinement user prompt share the same system prompt.
\subsection{System Prompt}
\begin{tcolorbox}[colback=gray!10, colframe=gray!50, breakable]
\small
You are an expert in causal inference and asia domain knowledge.
Your task is to generate or refine causal graph hypotheses representing causal relationships.
Always ensure the graph is a Directed Acyclic Graph (DAG) with no cycles.
\end{tcolorbox}
% \onecolumn
\subsection{User Prompt}
For initializing graph with LLM and experiments for LLM zero-shot capability, we use the prompt template for zero-shot generation, which generates a global graph based on variable names.

\begin{tcolorbox}[
    title=Prompt Template for Zero-shot Generation,
    colback=gray!5!white,
    colframe=gray!75!black,
    fonttitle=\bfseries,
    arc=1mm,
    breakable
]
\small
Generate an initial causal graph hypothesis for the \texttt{[Domain Name]} domain.

\textbf{Variables:} \\
\texttt{[Formatted Variable List]}
\texttt{[Domain Context]}\\
\textbf{Instructions:} \\
\textbf{THINK FIRST}: \\
1. Analyze the domain and relationships before proposing a structure\\
2. For each variable, determine its DIRECT CAUSES (parent variables)\\
3. Consider only direct causal relationships\\
4. Ensure the graph is a DAG (no cycles) - this is critical\\
5. Base reasoning on domain knowledge and temporal ordering\\

\textbf{Output Format (IMPORTANT - use this exact JSON structure):}

\begin{verbatim}
{
"reasoning": "Step-by-step reasoning about the 
causal structure before proposing the graph.
Explain your thought process, domain knowledge, 
and why certain relationships exist or don't 
exist.",
"nodes": [
{
    "name": "VariableName1",
    "parents": ["ParentVar1", "ParentVar2"],
},
{
    "name": "VariableName2",
    "parents": [],
},
]
}
\end{verbatim}

\textbf{CRITICAL:} 
\begin{itemize}
    \item Put ``reasoning'' FIRST before "nodes" to encourage thinking before outputting
    \item Use "nodes" as a LIST (not a dictionary)
    \item Each node must have "name" and "parents" fields
    \item "parents" must be a list (use [] for root nodes)
    \item Output ONLY valid JSON
\end{itemize}
\end{tcolorbox}

For step-by-step refinement, we utilize the prompt template for graph refinement. In this prompt, the model is asked to propose modification actions on edges.

\begin{tcolorbox}[
    title=Prompt Template for Graph Refinement,
    colback=gray!5!white,
    colframe=gray!75!black,
    fonttitle=\bfseries,
    arc=1mm,
    breakable
]
\small
Perform LOCAL amendments to the causal graph for the \texttt{[Domain Name]} domain.

\textbf{Variables:} \\
\texttt{[Formatted Variable List]}

\texttt{[Domain Context]}

\texttt{[Previous Reasoning Section]}

\texttt{[Causal Dossier (Confirmed Edges)]}

\textbf{Current Graph (Iteration [N]):} \\
\texttt{[Current Edges List]}

\textbf{Current Log-Likelihood:} \texttt{[Score]}\\

\textbf{ PREVIOUS FAILED ATTEMPTS \& FEEDBACK (CRITICAL: DO NOT REPEAT):}
\texttt{[memory\_content]} \\

% --- 记忆模块 2: 排除的操作 ---
% \begin{tcolorbox}[colback=red!5!white, colframe=red!50!black, size=small]
\textbf{ EXCLUDED Operations (previously failed, don't try again):} \\
\texttt{[ADD: ...; DELETE: ...; REVERSE: ...]}
% \end{tcolorbox}

\rule{\linewidth}{0.4pt}
\textbf{Refinement Instructions:}
\begin{enumerate}
    \item \textbf{Local Refinement}: Analyze the current graph and identify specific edges that could be added, removed, or reversed to better reflect the causal structure of the \texttt{[Domain Name]} domain.
    \item \textbf{Domain Knowledge}: Use your domain knowledge and the provided context to propose improvements. Focus on direct causal relationships, temporal ordering, and plausible mechanisms.
    \item \textbf{Strategic Selection}: Choose operations (ADD, DELETE, REVERSE) that you believe will most likely improve the graph's log-likelihood and accurately represent the system.
    \item \textbf{Update Dossier}: (Confirmed edges and tentative notes).
\end{enumerate}

\rule{\linewidth}{0.4pt}
\textbf{CRITICAL CONSTRAINTS:}
\begin{itemize}
    \item Propose at most \texttt{[num\_edge\_operations]} operations.
    \item Output ONLY valid JSON.
\end{itemize}

\textbf{Instructions:} \\
\textbf{THINK FIRST}: Before proposing operations:
\begin{enumerate}
    \item Analyze which changes would improve the likelihood based on the current structure and evidence.
    \item Choose operations that make sense based on domain knowledge and interventional data.
\end{enumerate}

\textbf{IMPORTANT - REVERSE operation semantics:}
\begin{itemize}
    \item If you want to reverse edge A$\to$B to B$\to$A, specify \texttt{"parent": "A", "child": "B"}
    \item The parent and child in your JSON should match the EXISTING edge direction
    \item Example: Current edge X1$\to$X2, to reverse it, use \texttt{\{"type": "REVERSE", "parent": "X1", "child": "X2"\}}
    \item This will change X1$\to$X2 into X2$\to$X1
\end{itemize}
\textbf{Output Format (IMPORTANT - use this exact JSON structure):}

\begin{verbatim}
{
"overall_reasoning": "...",
"confirmed_edges": ["VarA -> VarB"],
"edge_notes": {"VarX -> VarY": "Reasoning 
for keeping/tentative status"},
"operations": [
{
    "type": "ADD",
    "parent": "VariableName1",
    "child": "VariableName2",
    "reasoning": "Why this specific edge 
    should be added"
},
{
    "type": "DELETE",
    "parent": "VariableName3",
    "child": "VariableName4",
    "reasoning": "Why this specific edge 
    should be removed"
},
{
    "type": "REVERSE",
    "parent": "VariableName5",
    "child": "VariableName6",
    "reasoning": "Why edge VariableName5
    ->VariableName6 should be reversed to 
    VariableName6->VariableName5"
}
]
}
\end{verbatim}

\textbf{CRITICAL:} 
\begin{itemize}
    \item Put ``overall\_reasoning'' FIRST to encourage strategic thinking
    \item Output ONLY valid JSON
    \item Include UP TO 1 operations
    \item Valid operation types: ``ADD'', ``DELETE'', ``REVERSE''
    \item Each operation must have: type, parent, child, reasoning
    \item For REVERSE: always use the EXISTING edge direction (parent$\to$child) in your JSON
\end{itemize}
\end{tcolorbox}

\section{Case Study}
\begin{figure*}[htbp]
    \centering
    \includegraphics[width=0.8\linewidth]{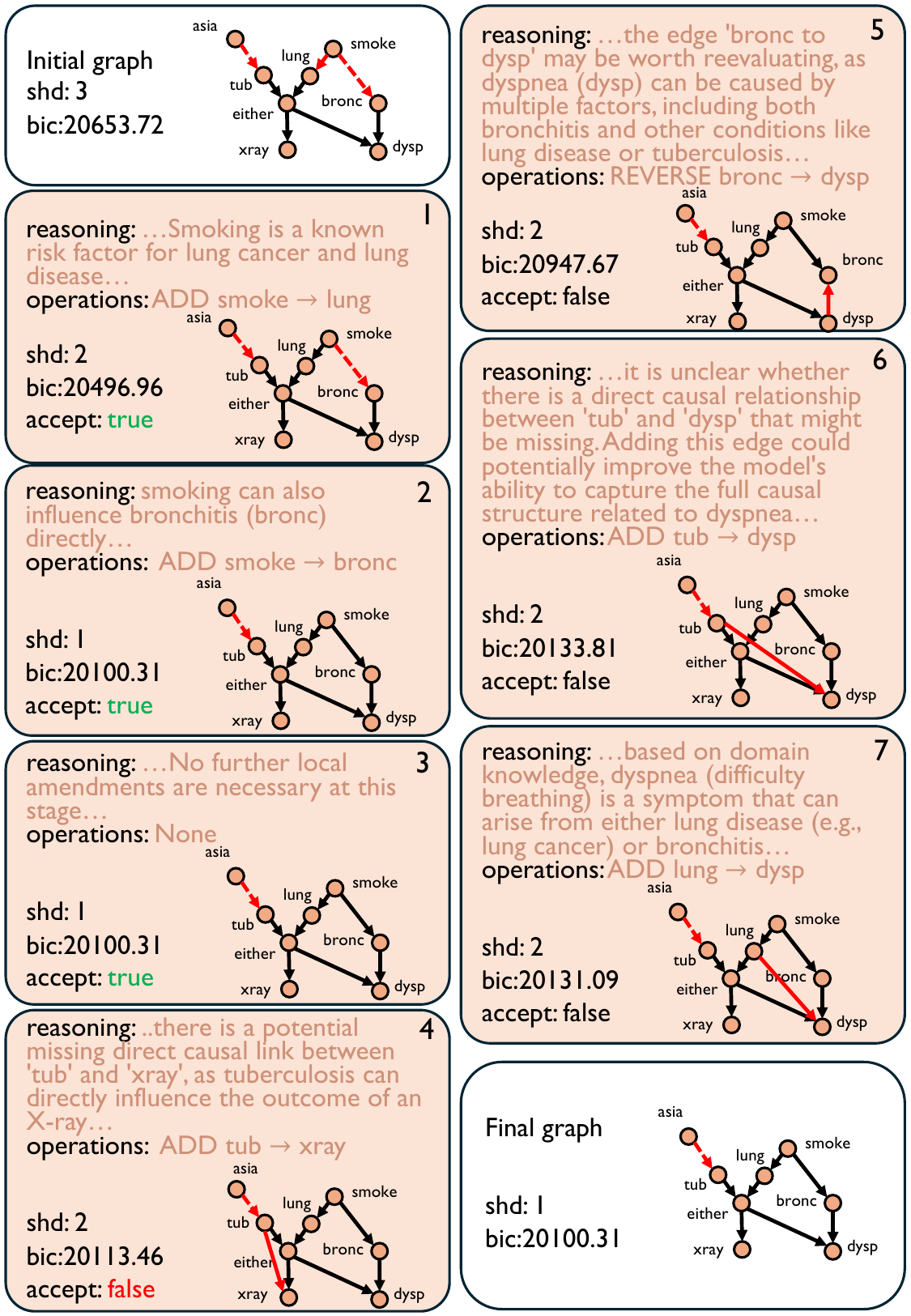}
    \caption{Case study on Asia. LLM conducts modifications on edges in 7 steps. We use Qwen3-14B in this case.}
    \label{fig:case}
\end{figure*}

Figure \ref{fig:case} illustrates a representative optimization trajectory on the Asia dataset, demonstrating the complementary roles of the LLM agent and the Statistical Verifier.

\begin{table*}[t]
    \centering
    \begin{tabular}{l|ccc|l}
        \toprule
        \textbf{Dataset} & \textbf{Nodes} ($d$) & \textbf{Edges} ($|E|$) & \textbf{Parameters} & \textbf{Domain} \\
         & & & & \\
        \midrule
        \textbf{Cancer} & 5 & 4 & 10 & Oncology \\
        \textbf{Asia} & 8 & 8 & 18 & Lung Diseases \\
        \textbf{Child} & 20 & 25 & 230 & Diagnosis \\
        \textbf{Alarm} & 37 & 46 & 509 & Monitoring \\
        \bottomrule
    \end{tabular}
    \caption{Detailed statistics of the benchmark datasets.}
    \label{tab:dataset_stats}
\end{table*}

\section{Dataset Details}
\label{app:dataset_details}

\subsection{Data Source and Distribution}
We evaluate our framework on four standard Bayesian Network benchmarks: \textbf{Cancer}, \textbf{Asia}, \textbf{Child}, and \textbf{Alarm}. The ground truth structures and parameters (Conditional Probability Tables) are obtained from the \textit{Bayesian Network Repository}\footnote{\url{https://www.bnlearn.com/bnrepository/}} provided by the \texttt{bnlearn} library \cite{scutari2010learning}.

\paragraph{Data Distribution.}
All variables are \textbf{discrete}. The data generation process samples directly from the ground-truth Conditional Probability Tables (CPTs). Consequently, the data follows a joint multinomial distribution strictly adhering to the underlying causal DAG.

\subsection{Experimental Setup}
\begin{itemize}
    \item \textbf{Sample Size:} Fixed at $N=5,000$ for all datasets.
    \item \textbf{Intervention Strategy:} We perform \textbf{perfect (hard) interventions}. We define $d+1$ distinct environments: one observational environment and $d$ interventional environments (where each node is intervened upon exactly once). In interventional samples, the target node is fixed to a random state, removing dependencies on its parents.
    \item \textbf{Sample Allocation:} To ensure a rigorous evaluation without biasing towards specific nodes, the sample budget is distributed \textbf{uniformly} across all environments ($N_{env} \approx \frac{N}{d+1}$). This setup creates a realistic "data-scarce" scenario for larger graphs (e.g., Alarm), where only $\sim 131$ samples are available per unique causal context.
\end{itemize}

\subsection{Dataset Statistics}
Table \ref{tab:dataset_stats} summarizes the benchmarks.

\subsection{Ground Truth Structures}
\label{app:ground_truth_viz}

\begin{figure*}[h]
    \centering
    \includegraphics[width=0.4\linewidth]{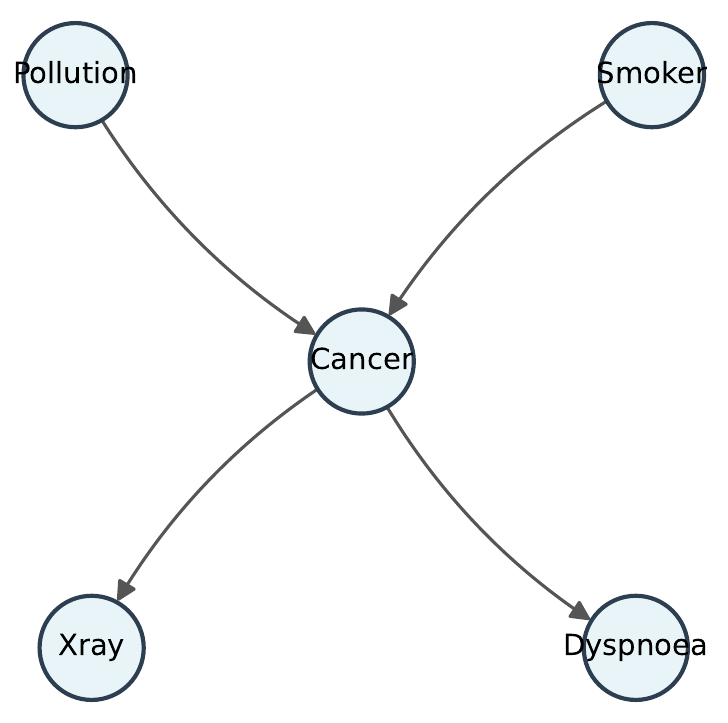}
    \caption{\textbf{Cancer} (5 nodes). A simple network representing oncological factors.}
    \label{fig:gt_cancer}
\end{figure*}

\begin{figure*}[h]
    \centering
    \includegraphics[width=0.6\linewidth]{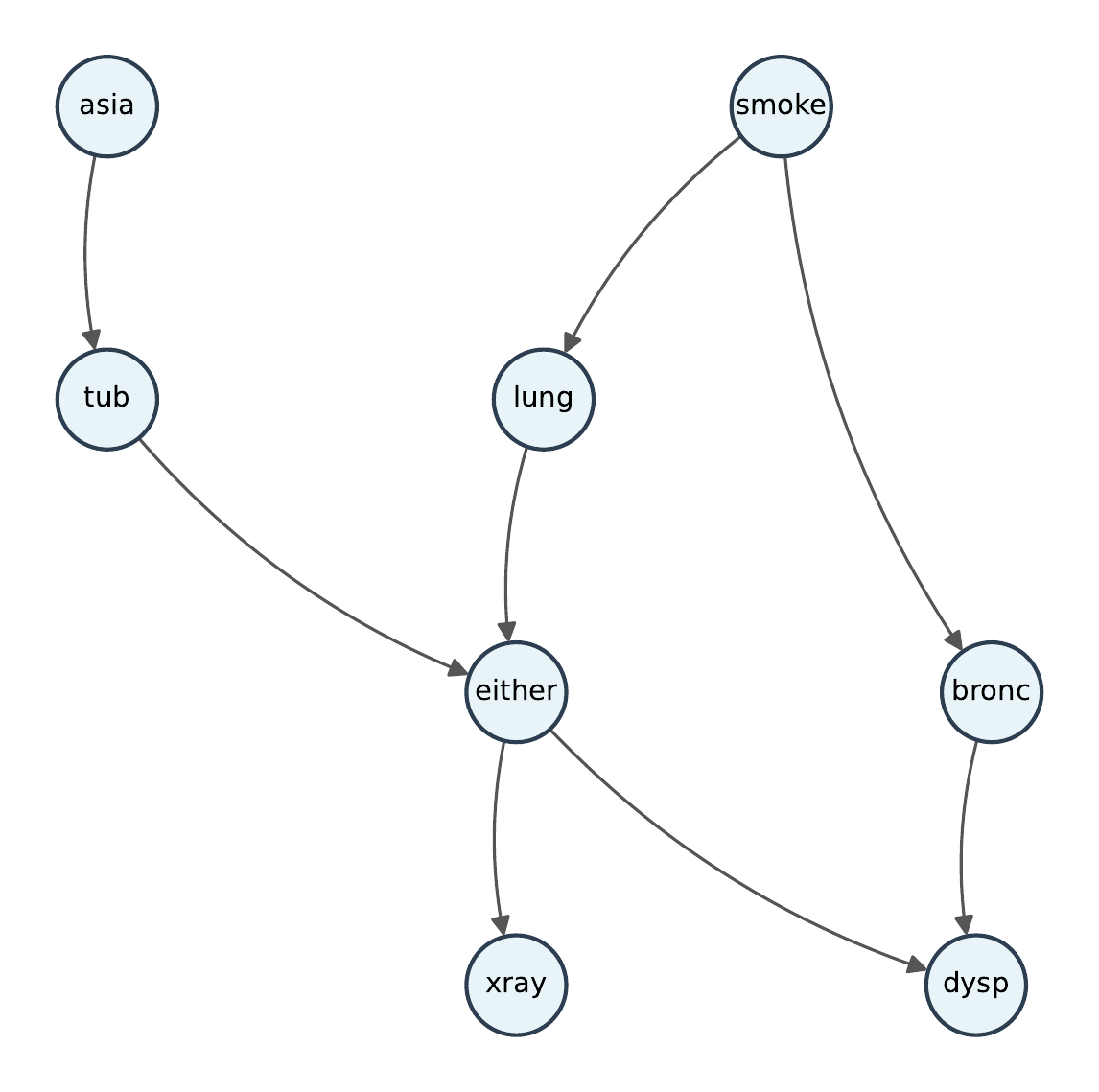}
    \caption{\textbf{Asia} (8 nodes). A moderate network involving lung diseases and travel history.}
    \label{fig:gt_asia}
\end{figure*}

\begin{figure*}[t]
    \centering
    \includegraphics[width=0.9\linewidth]{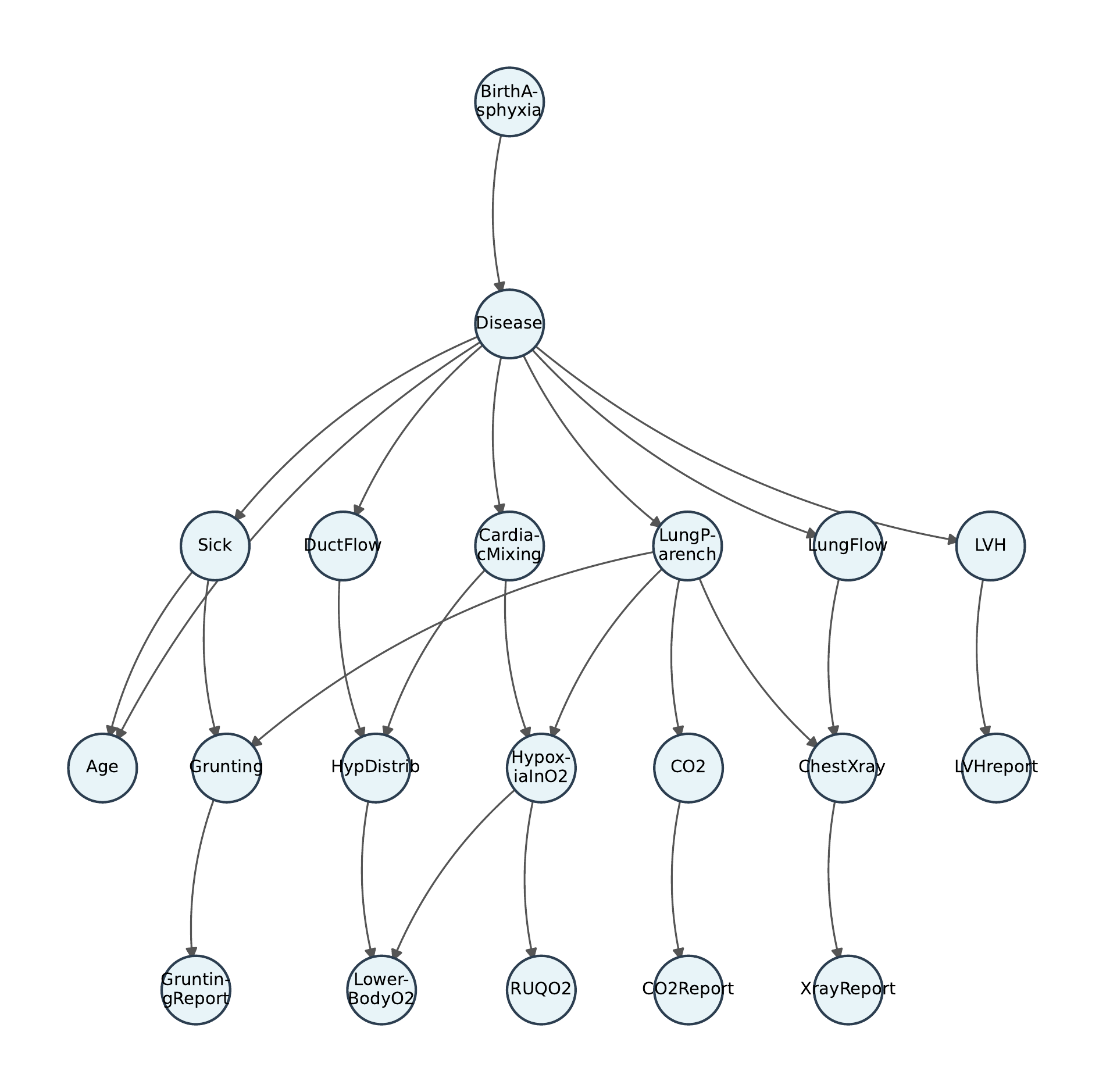}
    \caption{\textbf{Child} (20 nodes). A complex network for diagnosing congenital heart disease.}
    \label{fig:gt_child}
\end{figure*}

\begin{figure*}[t]
    \centering
    \includegraphics[width=1\linewidth]{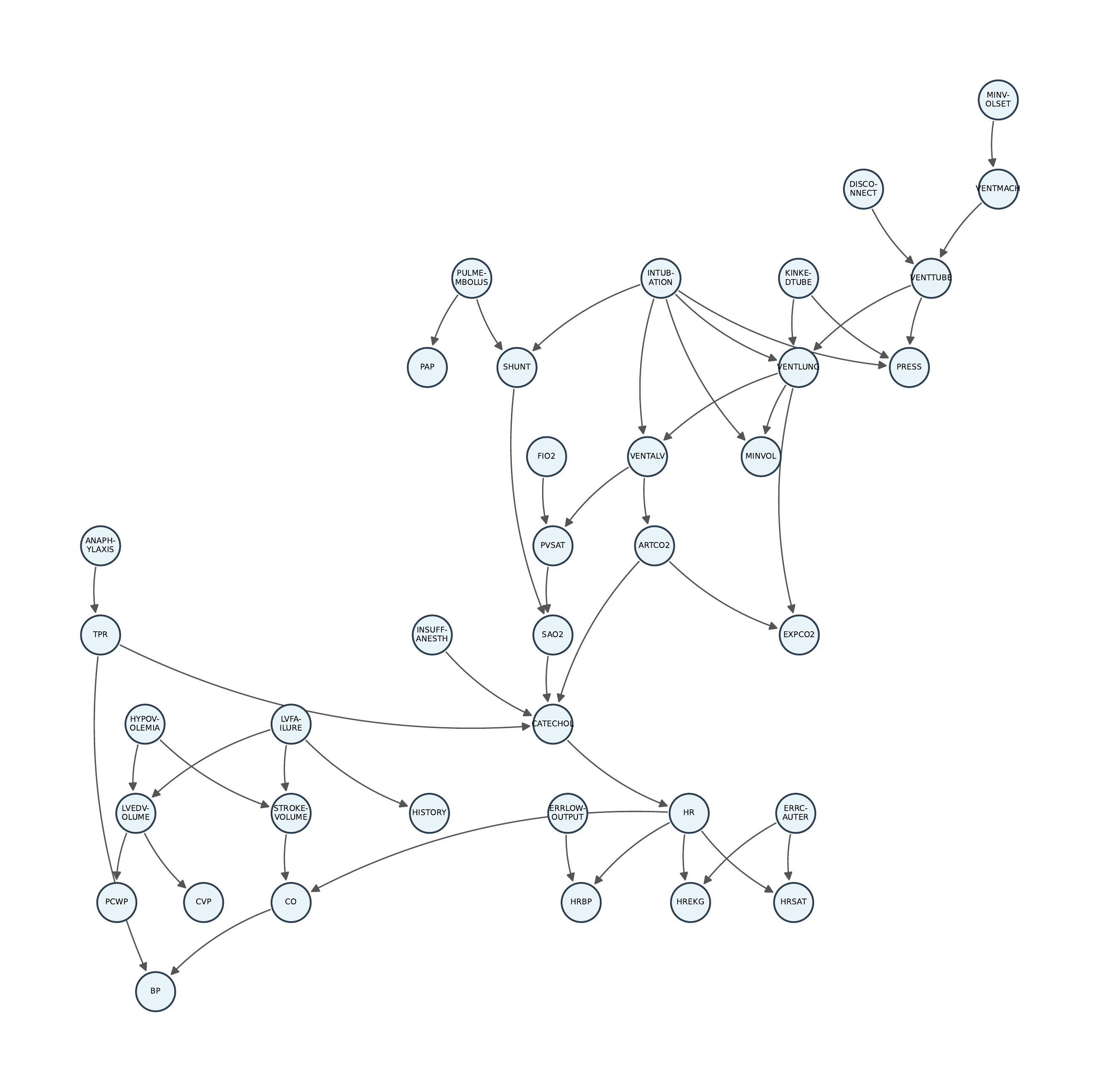}
    \caption{\textbf{Alarm} (37 nodes). A highly dense network for patient monitoring. The structural complexity presents significant challenges for pure statistical discovery.}
    \label{fig:gt_alarm}
\end{figure*}

To provide qualitative insight into the complexity of the causal discovery tasks, we visualize the ground truth Directed Acyclic Graphs (DAGs) for all four benchmarks. See Figure \ref{fig:gt_cancer},\ref{fig:gt_asia},\ref{fig:gt_child}, and \ref{fig:gt_alarm}.

\end{document}